\documentclass[a4paper,twoside]{article}

\usepackage{epsfig}
\usepackage{subcaption}
\usepackage{calc}
\usepackage{amssymb}
\usepackage{amstext}
\usepackage{amsmath}
\usepackage{amsthm}
\usepackage{multicol}
\usepackage{booktabs}
\usepackage{tikz}
\usepackage{pgfplots}
\usepackage{tabularx}
\usepackage{pslatex}
\usepackage{algorithm2e}
\usepackage{url}
\usepackage{multirow}
\usepackage{amssymb} 
\usepackage{pifont}

\usepackage{xcolor}

\usepackage{natbib}
\usepackage[bottom]{footmisc}
\usepackage{SCITEPRESS}
\definecolor{1}{HTML}{EFEFEF} 
\definecolor{2}{rgb}{0.1, 0.9, 0.9}

\begin{document}

\title{Visual Autoregressive Modelling for Monocular Depth Estimation}

\author{\authorname{Amir El-Ghoussani\sup{1}\orcidAuthor{0009-0009-5645-6684}, Andr\'e Kaup\sup{1}\orcidAuthor{0000-0002-0929-5074}, Nassir Navab\sup{2}\orcidAuthor{0000-0002-6032-5611}, Gustavo Carneiro\sup{3}\orcidAuthor{0000-0002-5571-6220} and \\Vasileios Belagiannis\sup{1}\orcidAuthor{0000-0003-0960-8453}}
\affiliation{\sup{1}Friedrich-Alexander University Erlangen-Nuremberg, Germany}
\affiliation{\sup{2}Technical University of Munich, Germany}
\affiliation{\sup{3}University of Surrey, United Kingdom}
\email{amir.el-ghoussani@fau.de}
}


\abstract{We propose a monocular depth estimation method based on visual autoregressive (VAR) priors, offering an alternative to diffusion-based approaches. Our method adapts a large-scale text-to-image VAR model and introduces a scale-wise conditional upsampling mechanism with classifier-free guidance. Our approach performs inference in ten fixed autoregressive stages, requiring only 74K synthetic samples for fine-tuning, and achieves competitive results. We report state-of-the-art performance in indoor benchmarks under constrained training conditions, and strong performance when applied to outdoor datasets. This work establishes autoregressive priors as a complementary family of geometry-aware generative models for depth estimation, highlighting advantages in data scalability, and adaptability to 3D vision tasks. Code available at "\url{https://github.com/AmirMaEl/VAR-Depth}".}

\onecolumn \maketitle \normalsize \setcounter{footnote}{0} \vfill

\section{Introduction}
Monocular depth estimation, a fundamental task in computer vision, involves predicting the depth of a scene from a single 2D image.
 At the same time, it is an inherently ill-posed problem. The process of capturing a 3D scene on a 2D sensor inevitably results in the loss of explicit depth information. 
 Inferring 3D structure from 2D projections presents several challenges. 
 First, scale ambiguity, where objects of varying sizes at different distances can produce identical 2D projections. Second, perspective ambiguity, as similar shapes at different orientations can result in indistinguishable 2D representations. Finally, occlusions can lead to incomplete depth information. Fortunately, deep neural networks partially address these challenges.

Earlier approaches to monocular depth estimation relied on supervised learning techniques~\citep{eigen_depth_2014,liuLearningDepthSingle2016,laina2016deeper}, which directly regressed depth from RGB inputs through convolutional neural networks. A significant paradigm shift occurred with the introduction of self-supervised methods~\citep{yin2018geonet,godard2017unsupervised,zhou2017unsupervised} that reduced the need for ground truth depth annotations by exploiting geometric constraints. These methods not only minimized annotation requirements but also allowed the use of vast, readily available datasets, significantly expanding the scope of training data in the field. More recently, advancements have been driven by the availability of large-scale training datasets~\citep{ranftl2020towards,ranftl2021vision}. 
This shift towards extensive data collection, combined with increasingly sophisticated neural network architectures, has enabled models to learn robust visual priors for depth prediction, at the cost of large training datasets. Despite their impressive results, these approaches contain significant computational overhead and extended training periods.

The latest advancements incorporate {generative models}~\citep{zhang2024vision} and adapt {large diffusion models}~\citep{ke2024repurposing,gui2024depthfm} for depth estimation, treating it as a conditional image generation task. However, these methods inherently rely on computationally expensive diffusion priors learned from massive datasets, often involving billions of image-text pairs~\citep{fu2025geowizard,gui2024depthfm}.

In this work, we explore {visual autoregressive (VAR)} modeling as an alternative to diffusion-based depth estimation. Although still slightly slower than a single feedforward pass, VAR models generate images hierarchically in a  few-step process, and can be pretrained on fewer samples while retaining strong generative priors. Building on Switti~\citep{voronov2024switti}, a large text-to-image VAR model trained on approximately 100 million image–text pairs, we adapt visual autoregression to affine-invariant depth estimation with three key design choices. 
First, we introduce a scale-wise conditional upsampling mechanism that propagates geometric cues across resolutions. Second, we design a guidance rule that balances contributions from the autoregressive prior and the conditional upsampler. Third, we propose a re-encoding strategy that enables intermediate predictions without requiring multiple upsamplers per scale. Fine-tuning relies solely on 74K synthetic images (Hypersim and vKITTI), keeping data requirements modest.

Our experiments demonstrate that VAR priors are a viable alternative to diffusion-based approaches. We report competitive indoor performance on NYUv2 and ScanNet, and reasonable outdoor results on KITTI, ETH3D, and DIODE. Beyond accuracy, our analysis compares sampling speed and the trade-offs associated with model size and training cost.

\section{Related work}
\subsection{Monocular Depth Estimation}
With the rapid development of deep neural networks, monocular depth estimation with the help of deep learning has been widely studied. Various supervised learning approaches have been proposed in recent years. The seminal paper of Eigen et al.~\citep{eigen_depth_2014} introduced a CNN-based approach to directly regress the depth. Liu et al.~\citep{liuLearningDepthSingle2016} propose utilising a fully-connected CRF as a post-processing step to refine monocular depth predictions. Other methods have extended the CNN-based network by changing the regression loss to a classification loss~\citep{wang2018occlusion,yin2018geonet,bhat2021adabins,laina2016deeper} or change the architecture of the depth estimation network entirely~\citep{liu2023vadepthnet,rudolph2022lightweight,richter}.
To address data scarcity and domain shift when training across different datasets, several works have explored domain adaptation techniques for monocular depth estimation, particularly leveraging synthetic-to-real transfer~\citep{kundu_adadepth_2018,el2025consistency}.

More recent advancement of depth estimation models has been marked by increasing model and dataset scales. MiDaS \citep{ranftl2020towards} established early benchmarks through multi-dataset training, followed by DPT \citep{ranftl2021vision} and Omnidata \citep{eftekhar2021omnidata} leveraging ViT architectures \citep{ranftl2021vision}. Recent works like DepthAnything \citep{yang2024depth} and Metric3D \citep{hu2024metric3d} further push boundaries by utilizing DINOv2 backbones \citep{oquab2023dinov2} and substantially larger datasets.

To reduce dependence on large training datasets, research has also explored the use of generative models, particularly diffusion models, for monocular depth estimation. DDP \citep{saxena2024surprising} introduces an encoder-decoder architecture that achieves state-of-the-art results on the KITTI benchmark \citep{geiger_vision_2013_kitti}. DepthGen \citep{saxena2023monocular} expands multi-task diffusion frameworks to include metric depth prediction, while DDVM \citep{saxena2023monocular} utilizes both synthetic and real data during pretraining to improve depth estimation performance.

A significant advancement in this field is Marigold \citep{ke2024repurposing}, which adapts Stable Diffusion~\citep{esser2021taming} for MDE, using only synthetic data during this finetuning process. However, it relies on a diffusion prior, in particular Marigold uses SD 2.0~\citep{rombach2022high} which is trained on LAION-5B~\citep{schuhmann2022laion}.
Instead, our method employs a generative VAR prior, i.e. Switti~\citep{voronov2024switti}, which was effectively trained using a significantly more constrained dataset of only 100 million samples.

To the best of our knowledge this work presents the first approach that is derived from a pre-trained text-to-image VAR model specifically designed for affine-invariant depth estimation.

\subsection{Autoregressive Modelling}

Autoregressive models have transformed the image generation task. Starting with PixelCNN~\citep{van2016conditional}, which predicted RGB values sequentially in raster scan order, the field evolved when \citep{van2017neural} showed images could be compressed into discrete tokens. VQGAN \citep{esser2021taming} later enhanced this approach by integrating adversarial and perceptual losses. 
Newer research focused on hierarchical and multi-scale approaches. VQVAE-2 \citep{razavi2019generating} used multi-level latent variables to capture both global and local features, while RQTransformer \citep{lee2022autoregressive} enhanced this with residual quantization. Parti \citep{yu2022scaling} scaled the ViT-VQGAN architecture to 20 billion parameters, achieving state-of-the-art text-to-image generation. MaskGIT \citep{chang2022maskgit} introduced masked prediction in VQ latent space, which was later adapted for video in MagViT models \citep{yu2023language,yu2022scaling}. MUSE \citep{chang2023muse} integrated the T5 language model \citep{raffel2020exploring}, strengthening text-image connections for better prompt responsiveness. More recent advances in autoregressive image generation have integrated advanced language modelling architectures. LLamaGen \citep{sun2024autoregressive} adapts the Llama foundation model \citep{Touvron2023LLaMAOA} for image synthesis, while AiM \citep{li2024scalable} incorporates Mamba's selective state space modeling \citep{gu2023mamba}. Lumina-mGPT \citep{liu2024lumina} extends this trend with a family of multimodal autoregressive models optimized for photorealistic image generation.

Visual Autoregressive Modelling (VAR)~\citep{tian2024visual} has recently achieved promising results in image generation by predicting entire resolution levels at once through next-scale prediction, thereby reducing the computational overhead while maintaining the high quality of the output. Several works have extended VAR to text-to-image generation: Switti \citep{voronov2024switti} implements text-image alignment via cross-attention modules. 
Similarly, Infinity \citep{han2024infinity} uses cross-attention for alignment while introducing an infinite-vocabulary optimizer and bitwise self-correction mechanisms to accelerate training. Infinity is trained on a combined dataset of LAION~\citep{schuhmann2021laion}, COYO~\citep{kakaobrain2022coyo-700m}. and OpenImages~\citep{kuznetsova2020open}. 
Building on these advances, we extend VAR beyond image synthesis, introducing a visual autoregressive prior explicitly adapted for monocular depth estimation. This establishes a new family of geometry priors, complementary to diffusion.

To the best of our knowledge this work presents the first approach that is derived from a pre-trained text-to-image VAR model specifically designed for affine-invariant depth estimation.

\section{Method}
Given the input RGB image $\mathbf{x} \in \mathbb{R}^{H \times W \times 3}$, our goal is to predict the depth map $\mathbf{d} \in \mathbb{R}^{H \times W\times 1}$ through an autoregressive process that decomposes the distribution $p(\mathbf{d}|\mathbf{x})$ into a sequence of conditional predictions. Inspired by VAR~\citep{tian2024visual}, we propose a multi-scale autoregressive approach to accelerate sequential depth prediction. 
The central component of our method is the integration of a prior, where we leverage pre-trained large-scale text-to-image VAR model and combine it with a custom conditional upsampling mechanism. This upsampling operates across the $k$ hierarchical VAR generation scales, progressively refining tokens from scale $k\rightarrow k+1$.

Our method is built on a hierarchical encoding and decoding scheme, which enables multi-scale representation and efficient autoregressive modelling (Sec.\ref{subsection:multi-token}). To effectively utilize the prior, we adopt a fine-tuning strategy that adapts the pre-trained VAR model to the depth estimation task~(Sec.~\ref{subsection:network}). Additionally, we design a conditional upsampling procedure that works with the prior, ensuring high-quality depth refinement during the upsampling process (Sec.\ref{subsection:training}). Finally, we detail the sampling process, in which we use a custom guidance strategy to combine both prior utilization and conditional upsampling~(Sec.\ref{subsection:guidance}).
\subsection{Background: Visual Autoregressive Modelling}
\label{subsection:multi-token}
VAR modelling~\citep{tian2024visual} establishes a hierarchical representation through a multi-scale VQGAN~\citep{esser2021taming} encoder, which transforms input images into a sequence of discrete token maps at progressively increasing resolutions. 
A transformer is then trained to autoregressively predict each higher resolution token map, conditioned on all preceding lower resolution maps.
\paragraph{Encoding and Codebook Lookup.}
Formally, during encoding, the image is transformed into $K$ token maps $R = (\mathbf{r}_1, \mathbf{r}_2, \ldots, \mathbf{r}_k)$, where $\mathbf{r}_1$ is the start token with $1\times 1$ image size and $\mathbf{r}_K$ corresponds to the token map with original encoded image size, i.e., $h \times w=h_K \times w_K$. The likelihood of each scale $k$ is factorized autoregressively across all scales $k$.

All token maps in $R$ share a common codebook $\mathcal{Z} \in \mathbb{R}^{V \times C}$, encoded via the encoder $\mathcal{E}$ and decoder $\mathcal{D}$. Here, $V$ denotes the vocabulary size and ${C}$ the number of channels. Initially, an input $\mathbf{x}$ is converted to an embedding feature map $\mathbf{f} = \mathcal{E}(\mathbf{x})$ with $\mathbf{f} \in \mathbb{ R}^{h\times w\times C}$, and subsequently interpolated to build $\mathbf{f}_k=\mathcal{E}_k(\mathbf{f}) \in \mathbb{R}^{h_k\times w_k \times C}$. $\mathcal{E}_k$ is defined as encoding the input with the encoder $\mathcal{E}(\cdot)$ with a subsequent interpolation to the current scale $k$. The feature map $\mathbf{f}_k$ is then mapped to a codebook entry $\mathbf{r}_k \in [V]^{h_k\times w_k}$ of the codebook $\mathcal{Z}$ via the closest Euclidean distance:
\begin{equation}
\label{eq:quantizer}
\mathbf{r}_k=\mathcal{Q}(\mathbf{f}_k)=\left( \underset{\mathbf{v} \in \mathcal{Z}}{\arg \min} ||\mathbf{v}-\mathbf{f}_k||_2 \right),
\end{equation}
where $\mathbf{v}$ represents one of the $C$-dimensional vectors from codebook $\mathcal{Z}$, and $\mathcal{Q}(\cdot)$ denotes the quantization operation.
The resulting multi-scale discrete tokens ${\mathbf{r}_1, \mathbf{r}_2, \ldots, \mathbf{r}_k}$ with $\mathbf{r}_k \in V^{h_k \times w_k}$ forms the foundation for autoregressive modelling.
\begin{equation}\label{eq:upsampling}
\hat{\mathbf{f}}=\sum_{k=1}^K \theta(\mathbf{f_k})=
\sum_{k=1}^K \theta(\mathbf{\mathcal{Q}^{-1}(\textbf{r}_k})).
\end{equation}
\paragraph{Decoding.}
The decoder $\mathcal{D}$ processes the combined feature representation $\mathbf{\hat{f}}$ to produce the final decoded output, such that $\mathcal{D}(\mathbf{\hat{f}}) = \mathbf{\hat{x}}$. To address information loss during upsampling, a network $\{\theta\}_{k=1}^K$ consisting of $K$ convolution layers is used.

We keep this standard formulation unchanged and focus on extending it with depth-specific conditioning and sampling strategies.

\begin{figure}
  \centering
    \includegraphics[scale=.25]{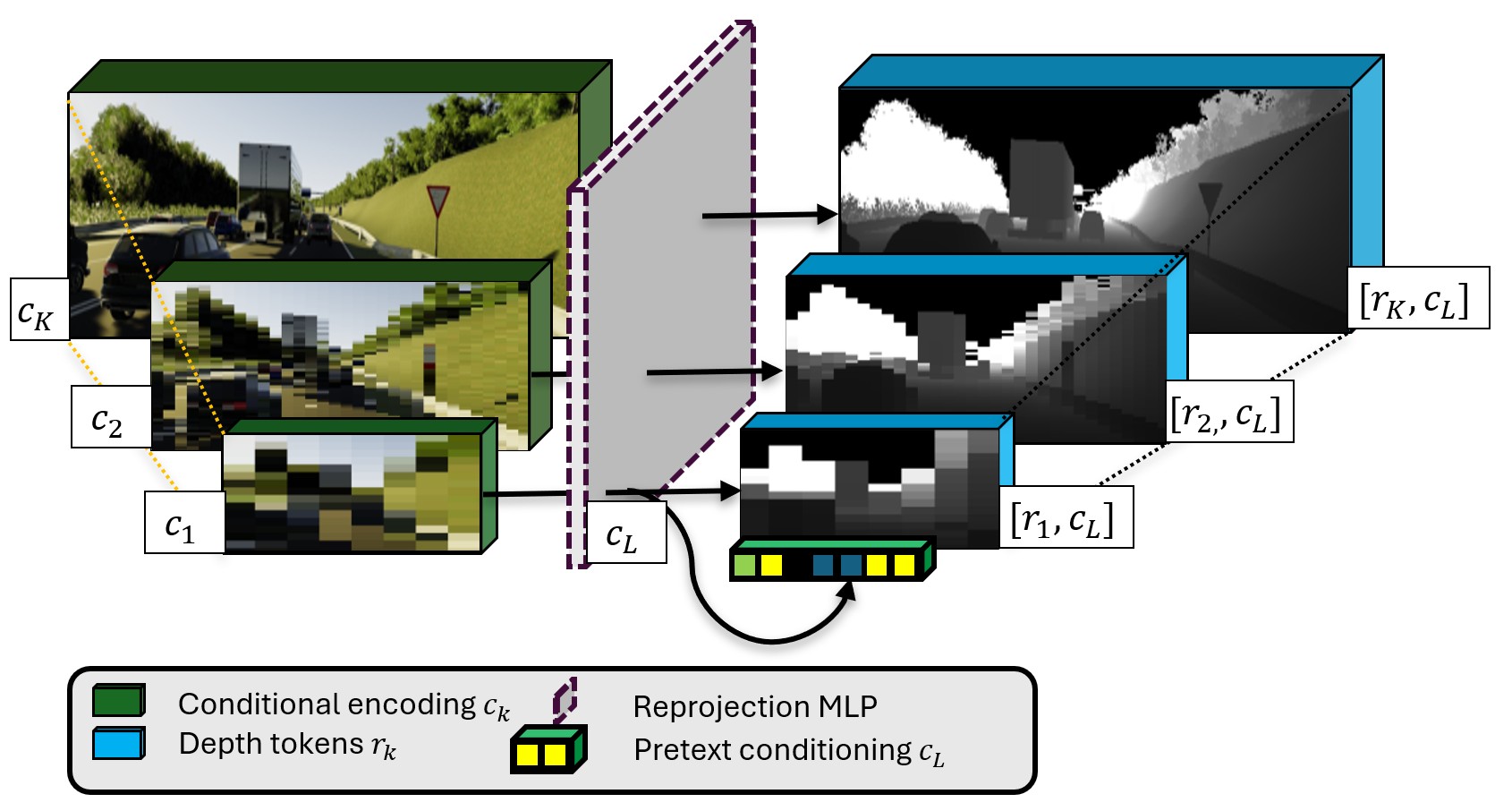}
    \caption{Overview of the proposed fine-tuning protocol. During model training  the RGB image $\mathbf{c}_k$ is reprojected using a MLP to align the dimension $\mathbf{c}_k$ with the text dimension $\mathbf{c}_L$. The modified input each scale $k$ to the VAR transformer then consists of both the ground truth depth token $\mathbf{r}_k$ and the conditional encoding $\mathbf{c}_L$. }
    \label{fig:cond_finetuning}
  \hfill
  \end{figure}
\begin{figure*}
  \centering
    \includegraphics[scale=.32]{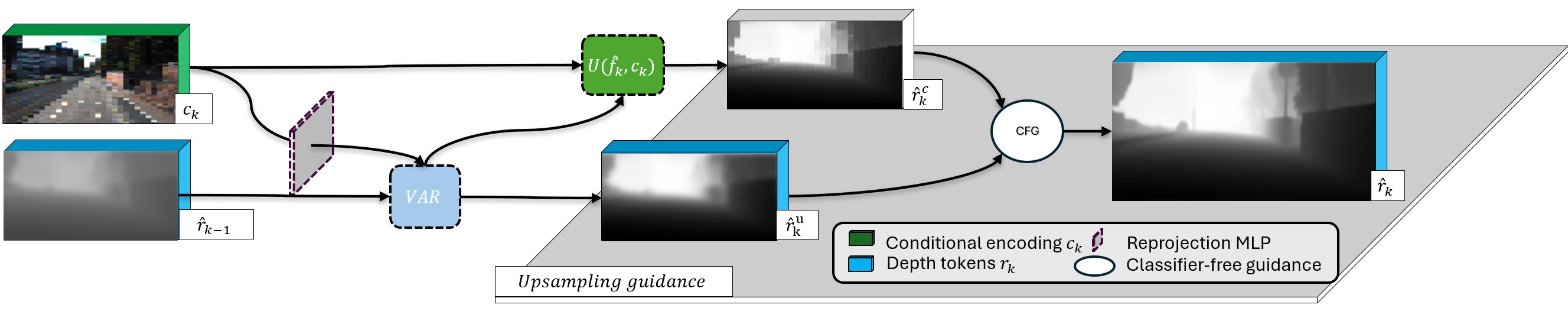}
    \caption{Overview of proposed sampling procedure. During sampling the prediction from the previous scale $\mathbf{\hat{r}}_{k-1}$ and the conditional reprojected encodings $\mathbf{c}_L$ are processed by the VAR transformer. After this processing step the VAR prediction $\mathbf{\hat{r}}^u_k=\mathbf{\hat{f}}_k$ is fed into the conditional upsampling network $U(\cdot)$, along with the condition encoding $\mathbf{c}_L$ to predict the output $\mathbf{\hat{r}}^c_k$. Finally both components are combined using classifier-free guidance to form the upsampled depth prediction $\mathbf{\hat{r}}_k$. }
    \label{fig:cond_upsampling}
  \hfill
  \end{figure*}

\subsection{Prior Model}
\label{subsection:network}
We adopt the pre-trained text-to-image model Switti~\citep{voronov2024switti}, which was trained in a two-stage process. Initial pretraining utilized 100 million high-quality, re-captioned image-text pairs filtered from a larger web dataset, progressing from 256x256 to 512x512 resolution. This was followed by supervised fine-tuning on approximately 40,000 manually selected, highly aesthetic and detailed text-image pairs at 1024x1024 resolution.

Our contribution in this stage is to extend Switti into a depth-conditioned VAR model by introducing a conditional fine-tuning phase, described in the following section.
\paragraph{Conditional fine-tuning.}
To adapt the pretrained VAR prior to depth estimation, we first replace the standard text-to-image encoder with a projection MLP $f(\cdot)$ that maps conditional encodings $\mathbf{c}$ into the text encoding space $L$:
\begin{equation}
\mathbf{c}_L = f(\mathbf{c}_k).
\end{equation}
The dimensionality of $\mathbf{c}_L$ is independent of the scale $\mathbf{c}_k$, and $\mathbf{c}_L$ is used as the conditioning signal for the transformer.

We then train the model using the teacher forcing paradigm established by~\citep{tian2024visual}, following a progressive resolution strategy—starting at $256\times256$, advancing to $512\times512$, and finally reaching $1024\times1024$. For conditional fine-tuning both the depth tokens $\mathbf{r}_k$ and the conditional RGB encodings $\mathbf{c}_L$ are jointly processed during training. Training optimizes next-token prediction, where the model is trained to predict target token distributions based on previously observed tokens and conditioning features. The loss function is defined as:

\begin{equation}
\mathcal{L}_{\text{TF}} = -\mathbb{E}_{\mathbf{r},\mathbf{c} \sim p_{\text{data}}} \left[\sum_{k=1}^K \log p_\theta(\mathbf{r}_k|\mathbf{r}{<k}, \mathbf{c}_{L})\right],
\end{equation}

where $\mathbf{r}_k$ represents tokens at the current scale $k$, $\mathbf{r}_{<k}$ denotes all tokens from previous scales, $\mathbf{c}{L}$ are the conditional encodings, $p{\text{data}}$ is the empirical data distribution, and $p_\theta$ is the model’s predicted probability distribution.

\begin{table*}[ht]
\centering
\caption{Results obtained for the evaluation on all datasets, NYU~\citep{nyu_depth}, KITTI~\citep{geiger_vision_2013_kitti}, ETH3D~\citep{Schops_2019_CVPR}, ScanNet~\citep{dai2017scannet} and DIODE~\citep{vasiljevic2019diode}. Baselines are sourced from Metric3D~\citep{hu2024metric3d}. Bold numbers represent the best performing method. We only compare to methods that use $<$ 3M datasamples for finetuning and use a prior trained with $\leq$ 100M samples. Best results are \textbf{bold}, second best results are \underline{underlined}.}
\label{tab:outdoor}
\tiny
\begin{tabular}{ccccccccccccccc}
\toprule
Method & Prior & training &  \multicolumn{2}{c}{NYU} & \multicolumn{2}{c}{KITTI} & \multicolumn{2}{c}{ETH3D} & \multicolumn{2}{c}{ScanNet} & \multicolumn{2}{c}{DIODE}\\
&& samples& AbsRel $\downarrow$& $\delta_1$ $\uparrow$& AbsRel$\downarrow$& $\delta_1$$\uparrow$ & AbsRel$\downarrow$ & $\delta_1$ $\uparrow$& AbsRel$\downarrow$ & $\delta_1$$\uparrow$ & AbsRel $\downarrow$& $\delta_1$$\uparrow$ \\
\midrule
DepthAnything v2 & DINOv2 - 142M & 63M&   4.4 &97.9 &7.5 &94.8 &13.2 &86.2 &— &— &6.5 &95.4\\
Metric3D & DINOv2 - 142M & 16M &   5.8 & 96.1 & 10.1 & 90.9 & 6.6 & 95.8 & 6.6& 95.0 & 31.7 & 77.2 \\

\midrule
Marigold & SD - 2.3B & 74K &   5.8 & 96.1 & 10.1 & 90.9 & 6.6 & 95.8 & 6.6& 95.0 & 31.7 & 77.2\\
Marigold v1.1 & SD - 2.3B &  74K &5.5 &96.4 &10.5 &90.2 &6.9 &95.7 &5.8 &96.3 &29.8& 78.2\\
DepthFM & SD - 2.3B & 63K & 6.5 &95.6 &8.3 &93.4 &— &— &— &— &22.5 &80.0 \\
Marigold E2E & SD - 2.3B & 74K &  5.2& 96.6 &9.6& 91.9& 6.2 &95.9 &5.8 &96.2 &30.2 &77.9 \\
\midrule \midrule
DiverseDepth & ImageNet - 1M & 320K & 11.7& 87.5 & 19.0 & 70.4 & 22.8 & 69.4 & 10.9 & 88.2 &  37.6    & 63.1 \\
MiDaS & ImageNet - 1M & 2M & 11.1 &88.5 &23.6& 63.0 &18.4 &75.2 &12.1 &84.6 &33.2 &71.5 \\
LeReS & ImageNet - 1M & 354K &9.0 &91.6& 14.9& 78.4& 17.1& 77.7& 9.1 &91.7& 27.1& 76.6 \\
HDN & ImageNet - 14M & 300K & 6.9&\textbf{94.8}&11.5&86.7& 12.1&83.3&8.0&\textbf{93.9}&24.6&\textbf{78.0} \\
DPT & ImageNet - 14M & 1.4M& 9.8 & 90.3 & \textbf{10.0} & \textbf{90.3} & \textbf{7.8} & \textbf{94.6} & \underline{8.2} & \underline{93.4} & \textbf{18.2} & {75.8} \\

\midrule

Ours - optimized $w_k$ & Switti - 100M & 74K & \textbf{6.4} & \textbf{94.8} &\underline{10.4} & \underline{90.1} & \underline{8.1} &  \underline{92.2} & \textbf{7.9} & \underline{93.4} & \underline{22.3} &{75.4}\\

\bottomrule
\end{tabular}
\end{table*}

\begin{table*}[ht]
\centering

\caption{Ablation on our sampling guidance.
We vary only the guidance schedule $w_k$ and report affine-invariant metrics (lower is better for AbsRel; higher is better for $\delta_1$). 
\emph{No guidance} ($w_k{=}-1$) disables our sampling contribution; 
\emph{Constant} uses $w_k{=}3.5$ for all scales; 
\emph{Optimized} is our schedule (guidance off for early scales, on for late scales).}
\label{abl:comp}
\footnotesize
\begin{tabular}{ccccc}
\toprule
Method  &  \multicolumn{2}{c}{NYU} & \multicolumn{2}{c}{KITTI} \\
& AbsRel $\downarrow$& $\delta_1$ $\uparrow$& AbsRel$\downarrow$& $\delta_1$$\uparrow$\\
\midrule
Ours - optimized $w_k$ & \textbf{6.4} & \textbf{94.8} &\textbf{10.4} & \textbf{90.1} \\
Ours - $w_k=-1\  \forall k$ (no guidance) & 12.4 & 55.7 & 20.1 & 70.5\\
Ours - $w_k=3.5 \ \forall k$ (constant guidance) & 9.2 & 85.1 & 12.1 & 84.5\\
\bottomrule
\end{tabular}
\end{table*}

\subsection{Conditional upsampling}
\label{subsection:training}
Conventional transformer conditioning appends conditional tokens to the input sequence. This ensures that predictions are based on both input and conditional tokens at each step. However, for complex geometric information (rather than simple textual prompts or classes), we have empirically observed that this signal does not propagate effectively through the Switti transformer across scales. This issue arises from the learned upsampling layer $\theta(\cdot)$ of Eq.~\ref{eq:upsampling}, which compresses or loses critical geometric context as information passes through generation scales $k$.

To overcome this limitation, we propose to replace the standard learned upsampling layer $\theta(\mathbf{f}_k)$, as shown in Eq.~\ref{eq:upsampling}, with a conditional upsampling layer $U(\mathbf{f}_k,\mathbf{c}_k)$ that operates after the geometric upsampling operation.  This conditional layer leverages an image-to-image model $U(\cdot)$ to predict the next token during the upscaling process:
\begin{equation}
\mathbf{\hat{f}}_{k+1} = U(\mathbf{\hat{f}}_k, \mathbf{c}_k).
\end{equation}
Here, $\mathbf{\hat{f}}_{k+1}$ represents the upsampled and predicted version of the previous token encoding $\mathbf{\hat{f}}_k$, upscaled to dimensions $h_{k+1}$ and $w_{k+1}$ using the upscaling network $U(\cdot)$. Notably, the network $U(\cdot)$ operates directly on continuous token encodings $\mathbf{f}$ and $\mathbf{c}_k$, rather than discrete tokenized representations $\mathbf{r}_k$ and $\mathbf{q}_k$.

We initialize $U(\cdot)$ with a U-Net~\citep{ronneberger_u-net_2015} using a timestep-aware approach, where the time step corresponds to step $k$. At each scale $k$, the model is trained to predict the final depth image $\mathbf{f}$ given the corresponding image encodings $\mathbf{c}_k$. During training, we minimize the $\mathcal{L}_2$-loss according to:
\begin{equation}
\mathcal{L}=||U_k(\mathbf{f}_{k-1},\mathbf{c}_{k-1})-\mathbf{f}||_2,
\end{equation}
where $U_k$ represents the image-to-image model at scale $k$. $\mathbf{f}_{k-1}$ is the ground truth token and $\mathbf{c}_{k-1}$ the RGB image condition encoding from the previous scale $k-1$.

This formulation enables the network to upsample the previous prediction $\mathbf{\hat{f}}_{k-1}$ to dimensions $h_k, w_k$, while effectively only requiring prediction of the residual ${s}_k=|\mathbf{f}-\mathbf{f}_k|$. $U_k(\cdot)$'s  lightweight ($\approx$40M parameter) design reduces fine-tuning complexity at the module level.,

\subsection{Sampling}
\label{subsection:guidance}
\paragraph{Upsampling guidance.}
Our goal is to leverage the prior information learned during the conditional fine-tuning stage~(Sec.~\ref{subsection:network}) while ensuring that conditions are properly propagated through the upsampling process~(Sec.~\ref{subsection:training}). To achieve this, we combine the conditional generation component $\hat{\mathbf{r}}^u_k$, obtained from the conditional fine-tuning stage, with the conditional component $\mathbf{\hat{r}}^c_k$, produced by the upsampling model $U(\cdot)$ in our conditional upsampling architecture. As seen in fig.~\ref{fig:cond_upsampling}, these two components are combined at each sampling stage to form the final prediction $\mathbf{\hat{r}}_k$ . This integration is achieved using a custom classifier-free guidance approach~\citep{ho2022classifier} that is defined as:
\begin{equation}
\label{eq:guidance}
\mathbf{\hat{r}}_k = (1+w_k) \mathbf{\hat{r}}_{k}^c - w_k \mathbf{\hat{r}}_{k}^u,
\end{equation}
where $w_k$ represents the classifier-free guidance scale applied at scale $k$, which controls the influence between both components $\mathbf{\hat{r}}^u_k$ and $\mathbf{\hat{r}}^c_k$ on the final output $\mathbf{\hat{r}}_k$. 

 \paragraph{Scale-wise conditional upsampling.}
 Intermediate predictions $\mathbf{\hat{r}}_k$ across all scales $k$ are required during the hierarchical generation process (see Eq.~\ref{eq:upsampling}). The straightforward approach to obtaining these intermediate predictions $\mathbf{\hat{r}}_k$ (where $k \neq K$) is to train separate models for each generation scale, with each model specifically designed to predict upscaled tokens at its corresponding resolution. However, this approach significantly increases computational and training complexity. 
 
 Instead, we present the more efficient re-encoding approach. Our conditional upsampling model $U(\cdot)$ is designed to predict the final high-resolution depth encoding $\mathbf{\hat{f}}$ without propagating the intermediate generation scales $k$. When a prediction at an intermediate scale $k$ is required, we generate the high-resolution prediction using the image-to-image model $U(\mathbf{\hat{f}}_k,\mathbf{c}_k)$, pass it through the VAR encoding scheme, thereby \textit{re-encoding} the high-resolution prediction as a forward pass, and extract the token representation at the desired scale $k$ as:
 \begin{equation}
      \mathbf{\hat{f}}_k = \mathcal{E}_k(\mathbf{\hat{f}}).
 \end{equation}

Although re-encoding introduces an extra forward pass, this avoids training separate upsamplers for each scale, simplifying the architecture and reducing fine-tuning cost.

\section{Experiments}
\label{sec:experiments}

We conduct experiments on five standard scale-invariant depth estimation benchmarks, results can be seen in Tab.~\ref{tab:outdoor} and are discussed in the following section. 
Additionally we present ablations indicating the performance when applying the proposed conditional fine-tuning in combination with the upsampling guidance and ablate on the sampling efficiency.

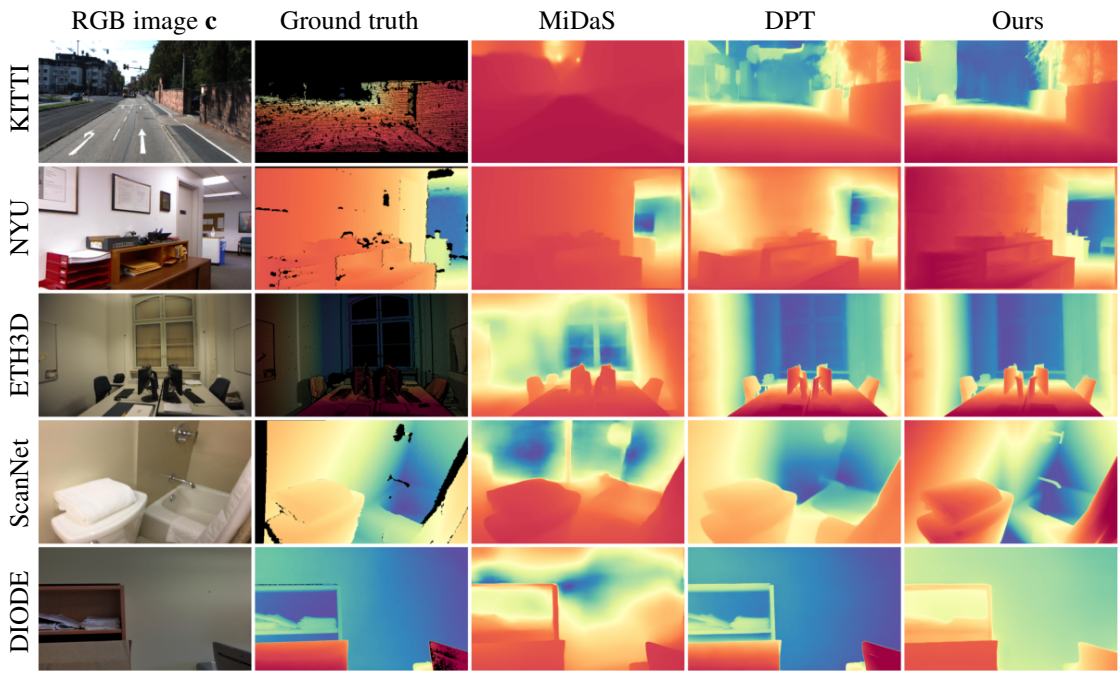
\begin{figure*}[h]
\centering

\begin{tikzpicture}

    \node at (0,0) {
        \includegraphics[scale=0.48]{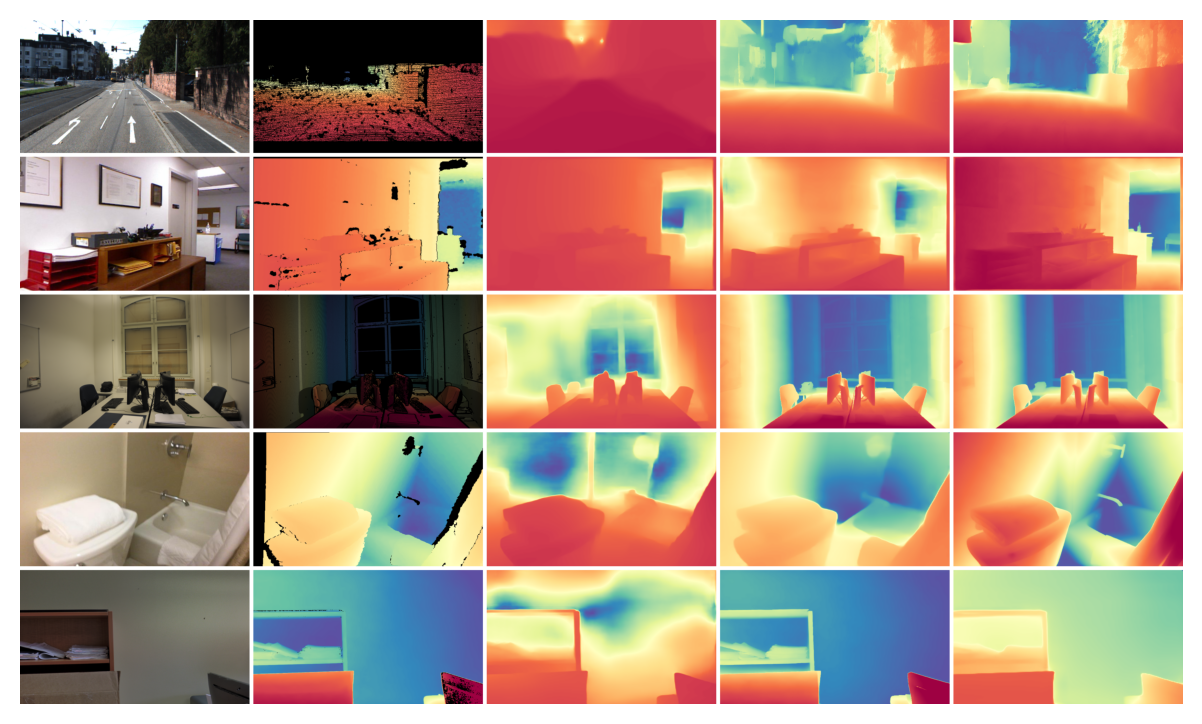}
    };
    \node[rotate=90, anchor=south] at (-7.1,0) {ETH3D};
        \node[rotate=90, anchor=south] at (-7.1,1.7) {NYU};
        \node[rotate=90, anchor=south] at (-7.1,-3.4) {DIODE};
        \node[rotate=90, anchor=south] at (-7.1,-1.7) {ScanNet};
            \node[rotate=90, anchor=south] at (-7.1,3.4) {KITTI};
            
    \node[anchor=south] at (-5.7,4.2) {\smash{RGB image $\mathbf{c}$}};
    \node[anchor=south] at (-3.0,4.2) {Ground truth};
    \node[anchor=south] at (0,4.2) {MiDaS};
    \node[anchor=south] at (2.8,4.2) {DPT};
    \node[anchor=south] at (5.8,4.2) {Ours};
    
\end{tikzpicture}
\caption{Predictions of Indoor and Outdoor experiments. Additionally we include the Ground truth and the input image provided by the test sets.}
\label{fig:short-a}
\end{figure*}
\subsection{Results}
\paragraph{Training datasets.}Following the existing  protocol~\citep{ke2024repurposing,gui2024depthfm}, we employ the Hypersim dataset~\citep{hypersim2021}, comprising 461 photorealistic indoor scenes. Images are standardized to $480\times 640$ pixels. In addition we utilize vKITTI~\citep{gaidon_vkitti_2016}, containing 20K additional samples cropped to KITTI benchmark specifications~\citep{geiger_vision_2013_kitti}. Depth values undergo normalization following~\citep{ke2024repurposing}.
\paragraph{Testing datasets.} The generalisation capability of the model is evaluated using five different real-world datasets. Each dataset represents a different environmental condition and acquisition process. First, the indoor domain is represented by NYU~\citep{nyu_depth}, comprising 654 RGB-D images from the designated test split, acquired with the Kinect sensor. Additionally ScanNet~\citep{dai2017scannet}, containing 1512 samples is utilized to further investigate the indoor prediction capability of our model.
The automotive domain is assessed through KITTI~\citep{geiger_vision_2013_kitti}, utilizing the Eigen test split configuration with LiDAR-based depth acquisition, representative of outdoor scenes. We additionally employ ETH3D~\citep{Schops_2019_CVPR}, consisting of 454 high-fidelity samples with multi-view stereo reconstruction ground truth, representing additional outdoor scenes. Finally the performance is evaluated on DIODE~\citep{vasiljevic2019diode} containing both indoor and outdoor scenes.

\paragraph{Comparison with other methods.} 
Table \ref{tab:outdoor} presents our depth estimation model's performance against baselines. Our objective is to demonstrate competitive performance under constrained data and compute, without relying on extremely large priors or extensive task-specific fine-tuning. Accordingly, our primary comparison set is limited to methods whose priors are trained on at most 100 million samples and that require fewer than 3 million samples for fine-tuning. These criteria define a practically relevant regime in which neither massive pretraining nor large-scale depth supervision dominates performance.

This setting is particularly relevant for applications such as robotics, embodied perception, and AR/VR, where collecting large annotated depth datasets is costly and rapid adaptation to new domains is preferred over retraining at scale.
Within this setting, our approach consistently outperforms nearly all baselines in indoor environments (NYU Eigen split, ScanNet validation split). In outdoor scenarios (KITTI Eigen split, ETH3D), our method delivers competitive performance, though slightly behind the strongest feedforward CNN baselines (DPT~\citep{ranftl2021vision}, HDN~\citep{zhang2022hierarchical}). We attribute this primarily to the domain gap: fine-tuning relies on Hypersim and vKITTI, which emphasize indoor and synthetic driving imagery. Despite this, our model achieves strong outdoor generalization given its modest fine-tuning set and smaller pretraining data. Overall, the performance gap remains narrow, highlighting the effectiveness of visual autoregressive priors in this constrained-data regime.

\paragraph{Implementation details.} Our implementation adopts the architectural foundation of VAR~\citep{tian2024visual} and the Switti transformer~\citep{voronov2024switti}. The autoregressive model adapts similar configuration to Switti, yielding a parameter count of approximately 2B. 
The optimization procedure employs the Adam optimizer~\citep{Kingma2014AdamAM} with a two-phase training protocol. Initial training is conducted on the synthetic vKITTI dataset for 30 epochs, followed by 80 epochs of training on the Hypersim dataset. The optimizer is configured with standard hyperparameters: momentum coefficients $\beta_1=0.9$ and $\beta_2=0.999$, numerical stability constant $\epsilon=10^{-8}$, and a learning rate of $\eta=10^{-6}$. The complete finetuning procedure requires approximately 100 NVIDIA A100 GPU hours.

The conditioning upsampling modules are trained independently using heavily augmented training data, which includes CutMix augmentations~\citep{yun_cutmix_2019} and ColorJitter augmentations. For optimization, the Adam optimizer~\citep{Kingma2014AdamAM} is employed with a learning rate of $\eta=10^{-4}$. The training strategy involves training the model for 10 epochs on vKITTI~\citep{gaidon_vkitti_2016} and 5 epochs on Hypersim~\citep{hypersim2021}. The total training time is approximately 5 hours.

\paragraph{Evaluation protocol. }Following the affine-invariant depth estimation paradigm~\citep{ranftl2020towards}, predictions $m$ are aligned to ground truth $d$ via least-squares fitting: $a=m \times s + t$. Performance is evaluated using two standard metrics~\citep{eigen_depth_2014,garg2016unsupervised,laina2016deeper}. We employ Absolute Relative Error (AbsRel) and $\delta_1$ accuracy. Computational efficiency is assessed through encoding and decoding latency measurements (in $ms$) for all evaluated approaches.

\begin{table*}[t]
\centering
\caption{Inference efficiency comparison. Latency measured on a single A100 GPU. The table emphasizes methods with iterative generative inference. Feedforward CNN baselines (e.g., HDN) are omitted when their inference behavior
is representative of other single-pass models. Our VAR approach achieves achieves inference latency comparable to reduced-step diffusion, while being slower and heavier than feedforward CNN baselines. The trade-off is that our method uses substantially fewer pretraining samples than diffusion.}
\label{tab:latency}
\footnotesize
\begin{tabular}{lcccccc}
\toprule
\textbf{Method} & \textbf{Prior Size} & \textbf{Fine-tune Samples} & \textbf{Inference Steps} & \textbf{Latency (ms)} & \textbf{GPUh} \\
\midrule
DPT & ImageNet -- 14M & 1.4M  & 1  & 45  & $>1000$ \\
DepthAnything v2  & DINOv2 -- 142M & 63M & 1 & 55  & $>1000$\footnotemark[1] \\
Metric3D v2  & DINOv2 -- 142M & 16M & 1  & 60  & $>1500$ \\
Marigold v1.0 & SD 2.0 -- 2.3B & 74K & $50\times 10$  & 4008 & $\sim$72 \\
Marigold v1.1 & SD 2.0 -- 2.3B & 74K & $1\times 4$  & 172 & $\sim$72 \\
Marigold E2E & SD 2.0 -- 2.3B & 74K & $1\times 1$  & 47 & $\sim$72 \\
Ours (VAR, opt. $w_k$) & Switti -- 100M & 74K & 10 (scales) & 53 & $\sim$100 \\
\bottomrule
\end{tabular}
\end{table*}
\paragraph{Qualitative results.}

Figure \ref{fig:short-a} presents depth prediction comparisons across different methods and datasets. As suggested by the quantitative measurements our approach performs very good on NYU and ScanNet, particularly our approach is able to capture the fine structures, such as armrest more reliably. On the KITTI outdoor dataset, our model demonstrates competitive performance with existing approaches. In particular when compared to MiDaS~\citep{ranftl2020towards} our approach is able to indicate richer details, e.g. tree structure. Finally both ETH3D and DIODE demonstrate that our approach is able to detect finer structures such as the window (middle row in Fig.~\ref{fig:short-a}), however finer details, such as the window frame are missed.

\subsection{Ablation studies}
\label{sub:ablations}
\paragraph{Sampling guidance.} We ablate our sampling contribution by varying the classifier-free guidance. $w_k=-1$ disables our proposed guidance (i.e., no conditional combination), reducing the model to a plain VAR prior conditioned only through teacher forcing. A constant schedule with $w_k=3.5$ applies strong guidance at all scales, while optimized $w_k$ enables guidance only at later scales $k\geq 6$. Table \ref{abl:comp} isolates this factor and shows that disabling guidance substantially degrades accuracy across datasets, whereas the optimized schedule yields the best overall results with minimal latency overhead. This confirms that the sampling/guidance mechanism is essential to the final model.

\paragraph{Efficiency.} We measure wall–clock inference latency on a single NVIDIA A100 (batch size 1) at the resolution $512 \times512$ in Tab.~\ref{tab:latency}. For feedforward baselines (DPT~\citep{ranftl2021vision}, DepthAnything~\citep{yang2024depth}, Metric3D~\citep{hu2024metric3d}), inference consists of a single forward pass. For Marigold, we report both typical multi-step denoising (50 steps) and reduced-step variants (1×1, 1×4)~\citep{ke2025marigold,martingarcia2024diffusione2eft}.  Our method requires 10 scale-wise stages, each predicting higher resolution maps, resulting in runtime similar to reduced-step diffusion.

Our method requires 10 scale-wise stages. Each stage predicts an entire resolution map, making runtime comparable to single-step diffusion. 
While our approach is somewhat slower than purely feedforward CNN baselines and the model has $\sim$2B parameters, it achieves this behavior with orders of magnitude fewer pretraining samples and comparable fine-tuning compute ($\approx 100$ GPU hours). This highlights a different efficiency–capacity trade-off, where VAR priors provide competitive inference while avoiding pretraining on extremely large datasets.

\section{Conclusion}
We presented adaptation of visual autoregressive priors for monocular depth estimation. By introducing a scale-wise conditional upsampling mechanism, classifier-free guidance, and a re-encoding strategy, our model achieves state-of-the-art indoor accuracy under constrained training data and competitive outdoor performance. Importantly, it requires 74K synthetic fine-tuning samples and performs inference via 10 fixed scale-wise steps, resulting in sampling speeds comparable to single-step diffusion and slightly slower than feedforward approaches.
While the parameter count remains, inference latency is comparable to diffusion-based methods despite using substantially less pretraining. This demonstrates that autoregressive priors achieve performance on par with diffusion while offering a complementary alternative with different trade-offs in training cost and inference structure. 
\footnotetext[1]{Given the $10^7–10^8$ image-scale pseudo-labeling/training, the total training compute is plausibly in the thousands to tens of thousands of GPU-hours range; exact figures are not reported.}
\section*{Acknowledgements}
Part of the research leading to these results is funded by the German Research Foundation (DFG) within the project 458972748. The authors would like to thank the foundation for the successful cooperation.

Additionally the authors gratefully acknowledge the scientific support and HPC resources provided by the Erlangen National High Performance Computing Center (NHR@FAU) of the Friedrich-Alexander-Universität Erlangen-Nürnberg (FAU). The hardware is funded by the German Research Foundation (DFG).

\bibliographystyle{apalike}
{\small
\bibliography{example}}

\end{document}